%% file: main.tex
\documentclass[9pt,conference,letterpaper]{IEEEtran}
\usepackage[T1]{fontenc}
\usepackage{cite}
\usepackage{amsmath,amssymb,graphicx,booktabs,dsfont,xcolor}
\usepackage{balance}
\usepackage[hidelinks]{hyperref}
\hypersetup{pdftitle={ChiProgramBench: Can LLM Agents Rebuild Efficient Circuits from Pin-Level Experiments?},pdfauthor={Tianrun Gou, Qingyao Yang, Wendong Xu, Liang Chen, Kuan Li, Ngai Wong},pdfsubject={arXiv preprint},pdfkeywords={hardware reconstruction, agents, verification, netlists, benchmark}}
\newcommand{\benchmark}{ChiProgramBench}

\input{analysis/task-scale-macros.tex}

\input{analysis/reward-intervals.tex}
\begin{document}
\bstctlcite{chiprogram:bibcontrol}
\title{\benchmark: Can LLM Agents Rebuild Efficient Circuits from Pin-Level Experiments?}
\author{\IEEEauthorblockN{Tianrun Gou\textsuperscript{1}, Qingyao Yang\textsuperscript{1,2},
Wendong Xu\textsuperscript{1,2}, 
Liang Chen\textsuperscript{1}, Kuan Li\textsuperscript{1}, Ngai Wong\textsuperscript{2}}
\IEEEauthorblockA{\textsuperscript{1}UniPat AI \qquad
\textsuperscript{2}The University of Hong Kong}}
\maketitle

\begin{abstract}
Understanding and rebuilding a system that can be run but not fully inspected is a core engineering
skill. One designs experiments, forms hypotheses about how the system works, tests and revises them, and finally builds a
replacement. To evaluate this capability of LLM
agents in the hardware domain, we present \benchmark{}, a benchmark of 18 tasks derived from
open-source processors,
accelerators, codecs, transforms, interconnects, and control circuits. Agents drive inputs, observe outputs, and submit a gate-level netlist for the replacement core using a public standard-cell library. The \emph{black-box} setting identifies clocks and resets, specifies the ports, capacity, and timing of any fixed SRAM outside the core, and lists the remaining pins by width and direction under anonymous names. The \emph{function-guided} setting also states the circuit's purpose and each pin's role. Hidden test cases compare the submitted netlist's outputs with the reference cycle by cycle. Power,
delay, and area earn credit relative to the reference only when every case
passes. Across 648
trials of 6 LLM agent configurations, successful agents recover algorithms
and numerical conventions, yet only 15.7\% of trials in each setting pass every case. The mean functional
score, the cycle-weighted share of passed cases, is 28.3\% in the black-box setting and 31.8\% with function guidance. Of the 102 fully passing designs, only 11 improve on the reference's combined power, area, and delay, demonstrating that behavioral recovery and
efficient implementation remain distinct challenges in hardware.
\end{abstract}


\section{Introduction}
Faced with a stateful system that can be run but not inspected, an engineer
works like an experimental scientist: design experiments, form hypotheses
about hidden state and timing, discard those the observations refute~\cite{moore1956gedanken}, and finally
build a replacement that is both correct and efficient. Recent agent
benchmarks treat such experimentation as a core
capability~\cite{arcagi3,jansen2024discoveryworld}. In hardware domain, this problem arises in practice when
third-party IP ships only as an encrypted simulation model~\cite{ieee1735} or
a legacy block outlives its design data. For any chosen input sequence, an engineer can compare the two circuits’ outputs cycle by cycle. The harder question is which sequence to try. A rebuilt image decoder, for example,
may decode an uninterrupted stream correctly but fail when its consumer stalls
or a reset interrupts a transaction. Agents must therefore choose input
sequences that expose such differences, then recover how computation, concurrent signals, and internal state interact
at clock boundaries.

Several benchmarks study individual steps in this process. Hardware design benchmarks ask AI agents to generate register-transfer-level (RTL) code, netlists, or physical layouts from explicit design specifications~\cite{liu2023verilogeval,lu2023rtllm,buildachip,lai2025analogcoder,analogdesignbench2026}, so agents
build without having to discover the behavior. Interactive benchmarks test the
experimenting step in abstract games~\cite{arcagi3} and simulated
laboratories~\cite{jansen2024discoveryworld}, but nothing has to be built from
what agents learn. ProgramBench brings together experimenting and implementation for software by asking an agent to rebuild
programs from an executable reference~\cite{yang2026programbench}. Compared with software rebuilding, hardware rebuilding must also
produce each output on the right clock cycle. Two designs that pass the same
tests can still differ in power, delay, and area.

We introduce \textbf{\benchmark{}}, a benchmark of 18 reconstruction tasks drawn
from open-source circuits in six domains. For each task, the reference circuit runs in its own container, separate from
the agent's sandbox; agents probe it across this boundary and submit a
replacement core mapped to the Nangate 45\,nm standard-cell library. The \emph{black-box} setting lists anonymized pins with their
directions and widths. It identifies clocks, resets, and any fixed SRAM interfaces. The
\emph{function-guided} setting also describes the circuit's purpose and each
pin's role. Hidden tests compare outputs cycle by cycle. Power, performance
(delay), and area (PPA) contribute to the reward only if the core passes every
test and meets the hold-time requirements.

\begin{figure}[!t]
\centering
\includegraphics[width=\columnwidth]{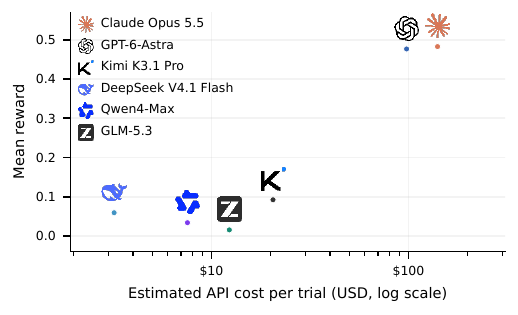}
\caption{Reward versus estimated API cost per trial, averaging both information settings equally. Costs use recorded token usage and public standard USD rates, including cached context and reasoning output. Each dot is a model's mean over 108 trials, with its logo just above. GPT-6-Astra runs in Codex, the others in Claude Code.}
\label{fig:cost-reward}
\end{figure}

Across 648 trials, the mean functional score (the cycle-weighted share of tests
passed) is 28.3\% in the black-box setting and 31.8\% with function guidance.
In both settings, 15.7\% of submissions pass every test. Of the 102 that pass,
11 improve on the reference power--area--delay product.
Figure~\ref{fig:cost-reward} compares the six agent configurations by mean
reward and estimated cost per trial.

\begin{figure*}[!t]
\centering
\includegraphics[width=\textwidth,trim=38 38 39 107,clip]{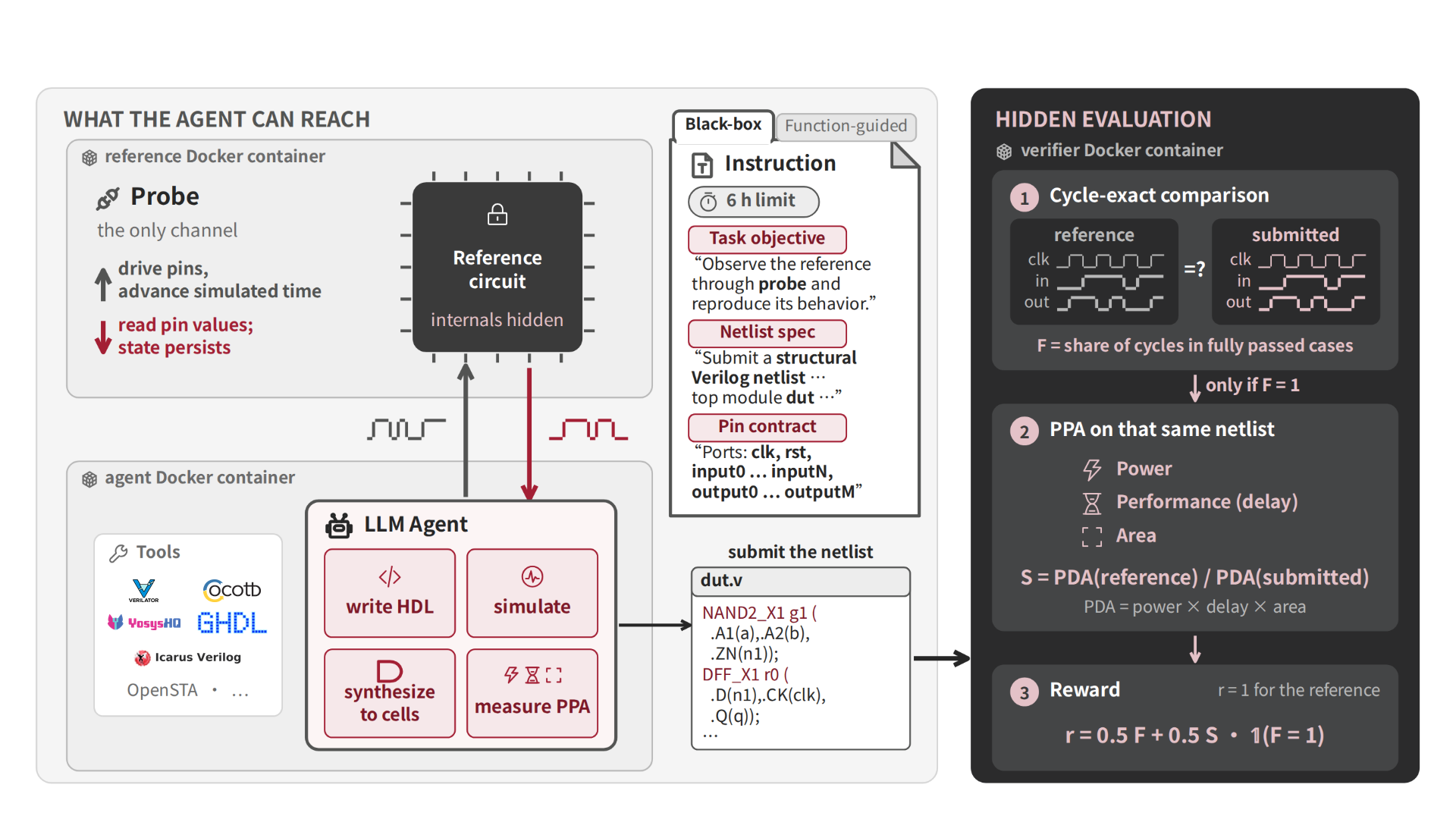}
\caption{Reconstruction and scoring. Agents probe a stateful reference running in a separate, network-isolated container and submit a mapped core netlist. Hidden tests measure functional recovery, with PPA credit enabled only after a full pass. The instruction tabs show the two settings.}
\label{fig:overview}
\end{figure*}

Our contributions are:
\begin{itemize}
\item \textbf{Reconstruction tasks with hidden cycle-exact tests.} A suite of
18 tasks in two information settings, with optimized gate-level references,
cycle-exact hidden tests, and faulty-design controls (broken netlists the tests
must reject).
\item \textbf{Cross-container probing and independent scoring.} Each reference
runs in its own network-isolated container, separate from the agent's sandbox,
and a stateful probe across the container boundary is the agents' only access
to it. A third, independent verifier container runs the hidden tests and links
cycle-exact functional testing to the PPA of the submitted core.
\item \textbf{Evaluation and Analysis of six agent configurations.} We compare circuit
correctness and hardware efficiency in the two settings. Case studies show how
agents used probe results to revise their designs and which errors their own
tests missed.
\end{itemize}

\section{Related Work}
\textbf{Hardware design from specifications.}
VerilogEval and RTLLM ask models to write RTL from a description of what the
circuit should do~\cite{liu2023verilogeval,lu2023rtllm}. The Comprehensive
Verilog Design Problems benchmark also tests debugging and
verification~\cite{pinckney2025cvdp}. RTLCoder trains models
for RTL generation~\cite{liu2025rtlcoder}, while AutoChip and RTLFixer use
simulation or compiler feedback to revise generated code~\cite{blocklove2025autochip,tsai2024rtlfixer}.
Other work gives agents electrical specifications for analog
circuits~\cite{lai2025analogcoder,analogdesignbench2026} or timing, power, and
area targets for digital designs~\cite{posteda,closer}. Build-A-Chip-Bench
starts from a workload specification and scores PPA after the design passes
functional and physical checks~\cite{buildachip}.

\textbf{Reconstructing software from observed behavior.}
CodeARC gives agents examples from a hidden Python function and lets them
query it on new inputs. Its two settings reveal or hide the function's
name~\cite{wei2025codearc}. ProgramBench gives agents a compiled program
and its documentation, then tests whether the codebase they write reproduces
the program's behavior~\cite{yang2026programbench}. Our function-guided
setting describes the circuit's purpose and each pin's role. Agents still
have to infer when each output should change.

\textbf{Learning hardware behavior from experiments.}
Active automata learning builds a finite-state model by querying a system
with sequences of inputs~\cite{angluin1987learning,vaandrager2017model}.
CacheQuery, for example, issues memory accesses to infer which line a
processor cache will evict~\cite{vila2020cachequery}. ProcReverser runs
agent-designed experiments to infer processor properties such as branch
prediction behavior~\cite{lyu2026procreverser}. In oracle-guided synthesis,
the algorithm chooses an input on which candidate programs disagree and uses
the reference's response to rule out candidates~\cite{jha2010oracle}. Other
methods learn combinational logic~\cite{rai2021logic} or a RISC-V
processor~\cite{cheng2024automated} from input--output examples.

\textbf{Recovering implementations from circuit structure.}
Researchers have used the wiring in a gate-level netlist to identify registers
and arithmetic blocks~\cite{albartus2020dana,wu2023gamoragraphlearningbased}.
Earlier work combined a partially known netlist with access to a working
chip to identify hidden keys or gate functions~\cite{subramanyan2015evaluating,elmassad2017reverse}.
LOCA uses a supplied netlist to recover formally equivalent RTL from sequential
designs~\cite{efe2026verified}. RTLREX starts from a combinational netlist
or a complete truth table, proves equivalence, and simplifies the resulting
RTL expressions~\cite{tang2026rtlrex}. In \benchmark{}, agents can observe
the reference only at its pins. We test the submitted standard-cell netlist
cycle by cycle and compare passing designs with the reference on power,
delay, and area.

\section{ChiProgramBench}

\begin{figure*}[!t]
\centering
\includegraphics[width=\textwidth]{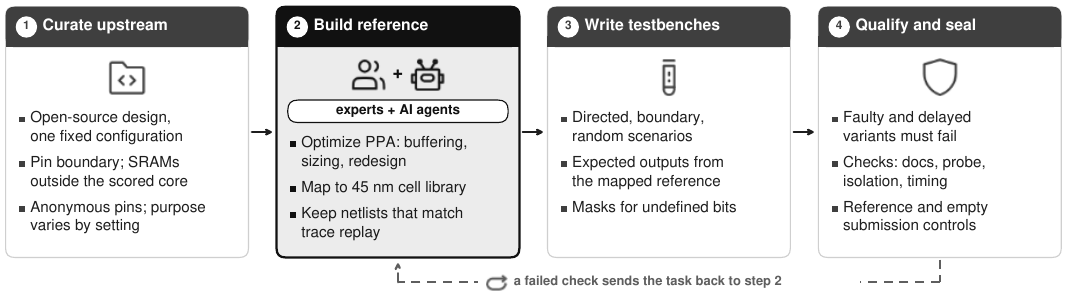}
\caption{Benchmark construction. Experts and AI-assisted development produce mapped references; workloads, isolation, and faulty-design controls are checked before release.}
\label{fig:construction}
\end{figure*}

Each task fixes a circuit configuration whose sequential behavior agents
reconstruct through the loop in Fig.~\ref{fig:overview}. Fig.~\ref{fig:construction} outlines
how tasks and references are built.
Across the 18 tasks, reference cores range from 3.7k to 648.4k cells, and
\TotalFunctionalCases{} hidden tests cover 37.3 million cycles (Table~\ref{tab:tasks}).
The NVDLA-derived task covers only
the convolution and postprocessing pipeline because the full design is too large to be verified with open-source EDA tools.

Tasks follow the Harbor format~\cite{harbor2026}: each package holds instructions, containerized environments, executable
verification tests, and our reference as its oracle solution, which also
normalizes agent PPA. 

\input{analysis/task-table-compact.tex}

\subsection{Reference Construction}
For each task, we curate an open-source design, define its interface and test
workloads, then build a verified mapped reference (Fig.~\ref{fig:construction}). Each reference starts as the mapped upstream RTL and is then optimized. We
restructure arithmetic (e.g., three-multiplier complex products and Booth or
Dadda trees), store delay lines and tables in latches, use one-hot AND-OR read
selection, and clock-gate idle registers. Synthesis tuning selects ABC~\cite{brayton2010abc} scripts and delay targets, keeps latch and clock-gate cells, buffers high-fanout nets,
and resizes cells according to timing slack. A revision is kept only if its measured power--area--delay product decreases
and it still matches the previous version on every defined output bit, in
gate-level replay of the workloads and in independent differential tests.

\subsection{Black-Box and Function-Guided Settings}
Both settings provide \texttt{instruction.md} and \texttt{OBSERVE.md}, specifying the submission interface, probe operations, output sample times,
scoring, and PPA measurement conditions, and both use the same per-task verifier and reference.
In both settings, clock and reset are named, and any fixed SRAMs are
documented by their ports, capacity, and access timing; other functional ports
appear
only as \texttt{input}$i$ or \texttt{output}$j$ with their widths, and
the \emph{black-box} setting gives nothing more. In the \emph{function-guided} setting, brief
descriptions add the circuit's purpose and each pin's role, such as operand,
result, valid, or request. During a trial, agents see their probe observations
and local tool results. The verifier runs the hidden tests after submission.

\subsection{Probing Across Separate Containers}
The reference and the agent run in separate containers, and agents observe the
reference only by probing it through a command-line tool and a Python library. A request lists inputs over time, either as timed
pin drives, clock advances, and reads, or as one row per cycle with fixed
drive and sample times; the probe simulates the reference and returns the
sampled outputs. The documentation states when outputs are sampled relative to
the clock and how unknown (\texttt{x}) bits are reported. The probe drives reset like any other input. Every stimulus used by the hidden tests can also be sent through
the probe.

Either request format can also run inside a session, which preserves circuit
state and simulated time across requests, allowing
the next stimulus to depend on the last observation; varying valid/ready timing
with fixed payloads, for example, can reveal acceptance and buffering behavior.
Closing and reopening creates a fresh instance. 

Each trial thus uses separate sandboxes. The reference container has no
network access and runs as an unprivileged user; the agent's container, also
without external network access, mounts only a read-only volume holding the
probe's Unix domain socket, the sole channel between the two, so reference
files never reach the agent. Hidden tests run after the trial in a third,
separate verifier container. For the cryptographic task \textit{Camellia}, we
also remove installed software implementations that could serve as
alternative references from the agent's container.

\begin{figure}[!t]
\centering
\includegraphics[width=\columnwidth]{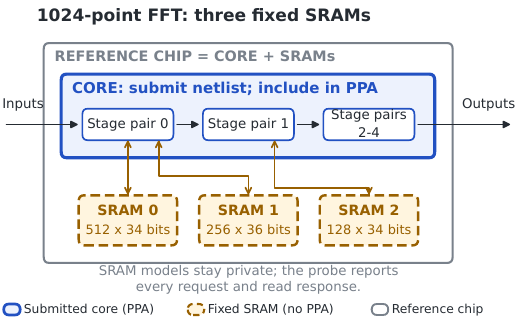}
\caption{Core boundary of the FFT task. The probed chip holds the core and three fixed SRAMs; agents submit only the core (blue), whose PPA is scored. Without fixed SRAMs, the core is the whole chip. The organization is schematic.}
\label{fig:boundaries}
\end{figure}

\subsection{Boundary of the Submitted Core}
Agents submit a gate-level netlist: structural Verilog that instantiates only
cells of the public Nangate45
library~\cite{nangate45,stine2007freepdk}, with top module \texttt{dut} and the
prescribed core interface. Multiple source
files and structural hierarchy are allowed. RTL, behavioral processes, delays, unsupported
cells, and initial-state overrides are rejected.
Figure~\ref{fig:boundaries} shows the core boundary of a task with fixed SRAMs.

Designated SRAMs stay fixed and outside the submission. Nangate45 supplies only
standard cells, and the open SRAM compiler OpenRAM~\cite{guthaus2016openram}
generates clocked macros with at most two ports, whereas the task memories
range from single-port SRAMs to a four-port array with combinational reads. Fixing these memories avoids evaluating large
arrays as artificial flip-flop implementations. Their interfaces, capacities,
masks, initialization, and access timing are prescribed and public, and the
probe reports every SRAM request and read response; the SRAM models stay in
the reference environment. Registers and buffers retained in the core count
toward PPA. Hidden tests replay reference memory responses and compare both functional outputs and memory requests. Candidates may change internal logic,
arithmetic, and state encoding, but must preserve pin-level behavior, fixed
memory capacities, and the timing of observable core--memory transactions.

\subsection{Hidden Tests and Scoring}
Hidden test cases cover computation,
reset and recovery, backpressure, arithmetic boundaries, and protocol
interactions. Each case is a fixed stimulus with prescribed output samples
over a fixed number of cycles; Table~\ref{tab:tasks} gives each task's case
count and total cycles. The verifier replays each case on the submitted
netlist in Icarus Verilog~\cite{icarus} with the Nangate45 cell models.
We do not prove formal equivalence with the reference: the hidden tests are
the specification, and a submission that passes every case is scored as fully
correct, even if untested inputs could reveal differences.

\textbf{Cycle-weighted Functional Passrate.} A case passes when every compared output bit
matches at every prescribed sample. A payload bit is compared only where the reference qualifies it: the associated valid,
request, or enable output is asserted, and the reference bit itself is 0 or 1. Those
qualifying controls are always compared. A candidate \texttt{x}/\texttt{z}
fails wherever the reference requires 0 or 1. Let $N_i$ be the fixed cycle
count of case $i$, and let $\pi_i=1$ indicate that the entire case passes:
\begin{equation}
F=\frac{\sum_i N_i\pi_i}{\sum_i N_i}. \label{eq:func}
\end{equation}

\textbf{Quality Score: Power, Delay and Area.} After a full pass ($F=1$), area $A$ sums
submitted leaf-cell areas. OpenSTA~\cite{opensta} computes power $P$ at the
library's typical corner from the candidate's own simulated activity over
fixed private workload windows, the same for candidate and reference, as total
window energy divided by total window duration. Our OpenSTA build, shared by
reference and candidates, adds one local patch: activity propagation also starts
at every VCD-annotated pin, so measured activity reaches logic outside
primary-input cones, such as free-running register loops; timing and power
formulas are unchanged. Input-driven initialization and
warm-up precede measurement. For power only, unknown startup values count as
zero and every measured cell pin must then stay binary, except in CV32E40P,
Ethernet, and OpenTPU, which keep four-state estimates; no default toggle rates are used.

OpenSTA also measures delay under fixed static-timing constraints. For capture-domain
period $T_j$, worst setup slack $s_j$, task reference period $T_{\mathrm{ref}}$, and
worst hold slack $H$ from an ideal-clock hold check on the same timing graph,
\begin{equation}
D=\begin{cases}
T_{\mathrm{ref}}\max_j\left[\max\left(0,1-\dfrac{s_j}{T_j}\right)\right], & H\ge 0,\\[2pt]
\infty, & H<0.
\end{cases}
\label{eq:delay}
\end{equation}
A hold violation thus removes all PPA credit. For one domain with
$T_{\mathrm{ref}}=T_j$ and $H\ge 0$, $D$ reduces to $\max(0,T_j-s_j)$. The library corner, clocks, I/O conditions, and constraints
are fixed per task. Clocks propagate through cell clock arcs, so clock-gating
cells are timed, and latch paths use OpenSTA's default time borrowing. SRAM
macros lie outside the timing graph: read data arrive 1\,ns after the clock
edge, and requests are captured at the next edge. In three tasks, the verifier also derives clocks generated by up to four
cascaded divide-by-two flip-flops. Any other unresolved clock,
a timing loop, an unconstrained endpoint, a design without constrained paths,
or a zero metric makes PPA unavailable; the verifier then reports a measurement
failure instead of a reward, which none of our full passes triggered. Power, delay, and the hold check use the library
wire-load model.

The PPA score $S$ and overall reward $r$ are\vspace{2pt}
\begin{align}
S&=\frac{P_{\mathrm{ref}}}{P}\frac{A_{\mathrm{ref}}}{A}
\frac{D_{\mathrm{ref}}}{D}, \label{eq:opt}\\[4pt]
r&=0.5F+0.5S\,\mathds{1}(F=1). \label{eq:reward}
\end{align}
The reference itself gives $S=r=1$, and improvements can exceed one.
Missing or invalid submissions receive zero. In all 36 task--setting
combinations, the complete verifier scores the reference at one and an empty submission at
zero, which in the Harbor framework is \texttt{oracle}${}=1$ and
\texttt{nop}${}=0$.

\textbf{Validation.} We also check consistency between probing and evaluation
through reference runs, probe replay, and session tests. Faulty-design controls
are structurally valid netlists with an output one cycle late, an inverted
address or output bit, a stuck write enable, or corrupted internal data. All controls fail at least one case ($F<1$) and so earn no PPA credit, including
mild variants meant to keep partial credit.
\section{Experiments}
\label{sec:experiments}

\begin{figure*}[!t]
\centering
\includegraphics[width=\textwidth]{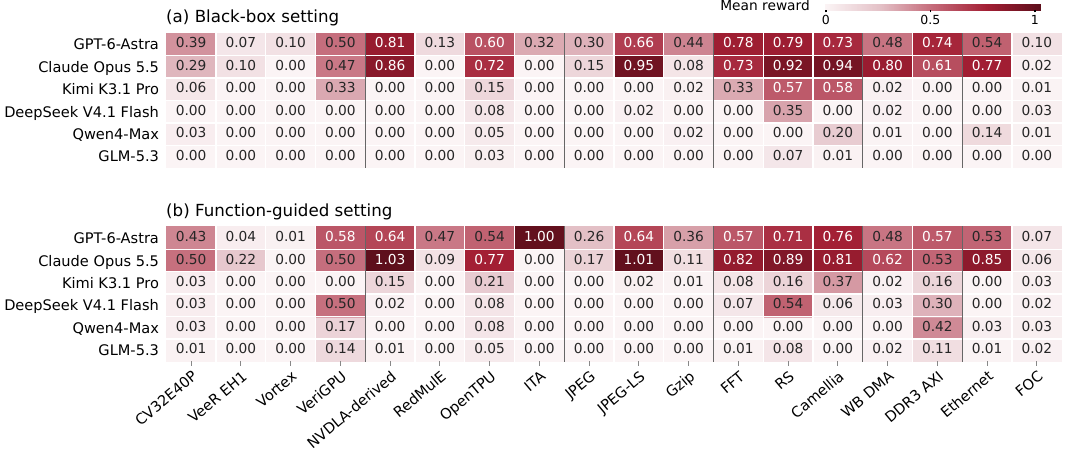}
\caption{Per-task mean rewards in the two information settings. Each cell averages three trials; both panels use the same color scale and task order as Table~\ref{tab:tasks}.}
\label{fig:settings-tasks}
\end{figure*}

\subsection{Setup}
We evaluate GPT-6-Astra with Codex CLI, and Claude Opus 5.5,
DeepSeek V4.1 Flash, Kimi K3.1 Pro, GLM-5.3, and Qwen4-Max with
Claude Code. Each model--scaffold configuration receives
three independent trials on every task in each setting, totaling 648 trials.
Trials start with public task files and an empty submission directory, use
maximum supported reasoning effort, and have a six-hour time
budget excluding setup and grading. Each agent container has 4--32 CPU cores and
8--64\,GB of memory, depending on the task, and no external network access;
agents reach the reference container only through the probe. The agent container provides
YosysHQ's OSS CAD Suite~\cite{osscadsuite}, including Yosys
0.70~\cite{wolf2013yosys} with ABC~\cite{brayton2010abc}, Icarus Verilog
14.0-devel~\cite{icarus}, Verilator,
GHDL, cocotb, Amaranth, and Migen, as well as Chisel~\cite{bachrach2012chisel},
SpinalHDL~\cite{spinalhdl}, Bluespec~\cite{bsc}, and Python. It also holds the
verifier's Nangate45 Liberty and cell models and its OpenSTA 3.1
build~\cite{opensta}, so agents can synthesize, simulate, and time candidate
netlists locally.

We average
the three trials within each task and weight the 18 tasks equally. All final
submissions contribute to the means.

\input{analysis/main-results.tex}

\begin{figure*}[!t]
\centering
\begin{minipage}[b]{0.49\textwidth}\centering\includegraphics[width=\linewidth]{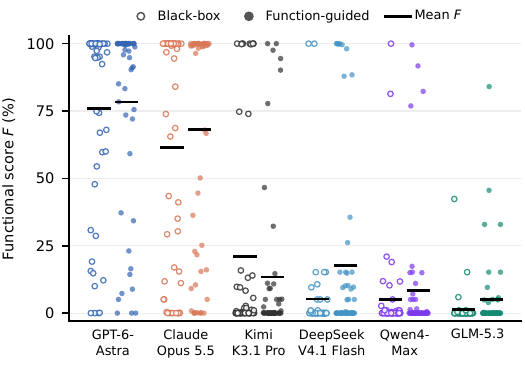}\\[1pt]{\footnotesize (a) All 648 trials}\end{minipage}\hfill
\begin{minipage}[b]{0.49\textwidth}\centering\includegraphics[width=\linewidth]{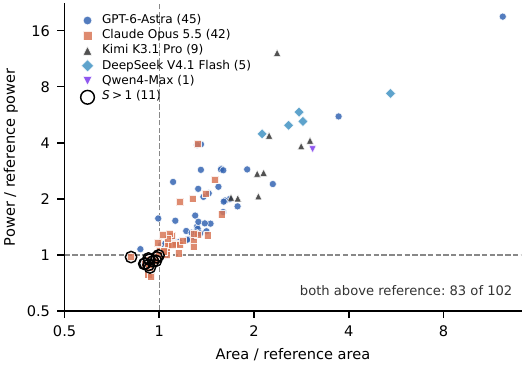}\\[1pt]{\footnotesize (b) The 102 full passes}\end{minipage}
\caption{(a) Functional score of all 648 trials, one column per model in its color: black-box trials as open markers (left) and function-guided trials as filled markers (right), 54 each, with horizontal jitter. Black bars mark the mean $F$. (b) Area and power of the 102 full passes relative to the reference (log scales). Dashed lines mark the reference; rings mark the 11 designs with $S>1$.}
\label{fig:trials}
\end{figure*}

\subsection{Overall Performance}
In each setting, GPT-6-Astra and Claude Opus 5.5 reach mean rewards of
0.47--0.50, while the other four configurations reach at most 0.115 in either setting
(Table~\ref{tab:main-results} and Fig.~\ref{fig:settings-tasks}) and stay below 0.10 when both settings are averaged
(Fig.~\ref{fig:cost-reward}). The mean reward difference (GPT-6-Astra minus Claude Opus 5.5) is $\BlackBoxModelDifference$ in the black-box setting and $\GuidedModelDifference$ with guidance. Both paired-task 95\% bootstrap intervals include zero. Seven tasks have no full pass in either setting. Of the 288 trials with $F=0$ in Fig.~\ref{fig:trials}a, 239 produced valid netlists for which the verifier recorded no passing case. The other 49 had a missing or invalid submission. Of the 258 partial passes, 55 reach $F\geq0.9$, including 25 above 0.99.

\begin{figure}[!t]
\centering
\includegraphics[width=\columnwidth]{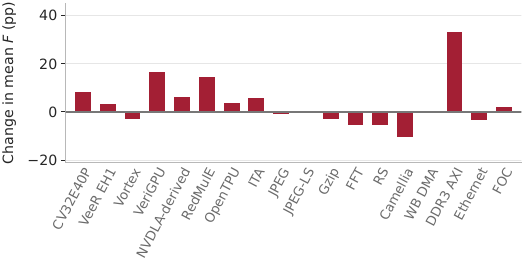}
\caption{Change in mean $F$ from the black-box to the function-guided setting for each task, averaged over the six configurations and three trials (18 trials per task and setting); pp denotes percentage points.}
\label{fig:context-gain}
\end{figure}

\subsection{Hidden-Test Results With and Without Function Guidance}
Mean $F$ increases by \GuidanceFunctionalMean{} percentage points with function guidance. A paired
bootstrap over the 18 tasks gives a 95\% interval of $[\GuidanceFunctionalLower,+\GuidanceFunctionalUpper]$ points
(10,000 resamples, keeping all trials of each sampled task together).
Three tasks account for the aggregate gain (Fig.~\ref{fig:context-gain}).
DDR3 AXI gains 33.2 points, VeriGPU 16.5, and RedMulE 14.3. With equal task
weights, the three together contribute \GuidanceFunctionalTopThree{} points to the overall mean. Across
the other 15 tasks, gains and losses average $\GuidanceFunctionalOtherMean$ points.
For each task, we also counted black-box trials in which the agent named
the circuit type and acted on that inference (Section~\ref{sec:capability-limits}).
Tasks with fewer such trials tended to gain more in mean $F$ from guidance
(Spearman $\rho=-0.62$ across the 18 tasks).

Each setting yielded 51 full passes in 324 trials (15.7\%). In the black-box setting,
those passes covered nine tasks. With function guidance, they covered
those nine plus ITA and VeriGPU. GPT-6-Astra accounted for all five full
passes on the two additional tasks, three on ITA and two on VeriGPU.
Its total rose from 20 to 25, while Kimi K3.1 Pro's fell from seven to two.
On Ethernet, GPT-6-Astra's mean $F$ fell from 99.8\% to 91.1\% with
guidance. Section~\ref{sec:guided-experiments} examines one reply that
the model recovered in a guided trial.

\subsection{PPA of Passing Netlists Relative to the Reference}
Fig.~\ref{fig:trials}b compares each full pass with the reference. Area and power both exceed the reference in 83 of the 102 full passes (median ratios 1.29 and 1.33), while delay stays close (median 1.01, at most 2.00$\times$). Only 11 designs improve the joint power--area--delay product ($S>1$): ten Claude Opus 5.5 designs (RS four times, JPEG-LS three times,
NVDLA-derived twice, Ethernet once) and one GPT-6-Astra ITA design.

\subsection{Repeatability on the Same Task}
For each model, task, and information setting, we compared three independent
trials. Of the 216 groups, 43 produced at least one submission that passed
every hidden case. Within those 43 groups, all three trials passed in 23,
exactly two passed in 13, and one passed in seven. For example, Kimi K3.1
Pro's three black-box FFT netlists passed 0, 101, and 83 of the same 101
cases, in trial order. In another 63 groups, all three trials scored $F=0$.

\section{Reconstruction Behavior Analysis}
\label{sec:reconstruction-analysis}

One black-box FFT trial shows how an agent identified a circuit from its
pins, implemented its behavior in RTL, and tested a design that passed
every hidden case. Across trials, we also see where these steps can fail.
Some agents never identified the circuit, some built designs that never
signaled a result, and some failed on inputs they had never tried.

\subsection{A Successful Reconstruction Trajectory}
GPT-6-Astra held the input constant and observed one nonzero result per
1,024 samples. SRAM reads ran three addresses ahead of writes, and the
values at the first two memory interfaces showed sums and differences.
These observations led the agent to infer a streaming FFT. Mode changes
revealed different scaling, and arithmetic probes identified 16-bit fractional
coefficients with a ties-to-odd rounding rule.

It also found that a burst of at least 512 samples launches a full output
frame and that one output flags interrupted frames; its notes record free-running
SRAM address pointers, frame control that counts valid samples, and a reset that
clears control state but retains data. A cycle-accurate model, RTL replay,
technology mapping, and gate-level checks followed. The agent then tried a
three-multiplier complex product, Booth multipliers, and latch-based delay
lines, timing each candidate under the wire-load model. The final core passes all 101 hidden
cases with $S=0.729$; its delay is close to the reference, but area and power
remain 15.8\% and 17.3\% higher.

\subsection{Capability Limits}
\label{sec:capability-limits}
\textbf{Recognizing the circuit.} In 183 of 324 black-box trials (56.5\%),
the agent named the specific circuit type and used that inference in its work.
Another 24 identified only a broader class. On the NVDLA-derived convolution
task, for example, the most specific identification was a dot-product pipeline.
All 18 Camellia trials recognized the cipher. In one, GPT-6-Astra matched the
output for an all-zero block to a known test vector. All 51 black-box full
passes came from trials that recognized at least the class. Only 44 of the 183
exact identifications led to a full pass.

\textbf{Implementing the core function.} In one black-box trial, Claude Opus 5.5
recognized ITA as an attention accelerator and failed all 138 cases. Across its
six trials, Opus passed at most one case per trial. Two submissions held the
result-valid signal low, so the receiver could never accept a result. Two
function-guided submissions recovered some numerical results but still missed
the timing of transfers and SRAM accesses. GPT-6-Astra's three black-box
designs passed 133, 137, and 1 of the 138 cases. All three of its
function-guided designs passed the whole suite, including the checks on when
results and memory requests appeared at the pins.

\textbf{Probing edge cases.} Claude Opus 5.5 passed all 169 Ethernet cases in
five of six trials. In the remaining black-box trial, it tested addresses near
a subnet boundary and revised its circuit to broadcast only within that subnet.
Its probes missed the all-ones IPv4 address (\texttt{255.255.255.255}). The
reference sends packets for this address directly as local broadcasts. The
submitted circuit treated the destination as remote and tried to look up the
gateway's Ethernet address. All 25 failed cases first diverged at the same pin.
The reference marked its IP transmitter busy while the submission left the
signal low.

\subsection{Choosing Probe Inputs from Port Descriptions}
\label{sec:guided-experiments}
On Ethernet, the function-guided description names the Address Resolution
Protocol (ARP) and labels the ports carrying packet headers and payloads.
In one such run, GPT-6-Astra tried 12 ARP operation codes. It found that the
reference returned its IP address when an inverse ARP request named the
chip's own hardware address as the target. The agent added this condition to
its RTL and passed all 169 hidden tests. In an independent black-box run,
the same model never tried this request. Its circuit passed 167 tests and
remained silent in the two that required this reply. Across the three trials per
setting, one black-box submission and two function-guided submissions passed
both reply tests.

On the NVDLA-derived task, the port descriptions identified the controls
that bypass arithmetic stages. After its fixed-control tests matched the
probe, GPT-6-Astra toggled those controls while data were in flight.
It observed that a bypassed stage still drained the data already in its queue.

\section{Discussion and Conclusion}
Agents recover complex computations and subtle timing rules from pin-level
experiments, yet full passes remain rare and most passing designs cost more
than the reference. The public probe can express every input sequence used by
the hidden tests. During their six hours, agents had to decide
which sequences to try. On Ethernet, testing an inverse ARP request exposed
a reply condition that another run missed.
Isolating the benefit of choosing probes adaptively would require a comparison
with fixed input sequences at the same observation budget.

Because PPA counts only after a full pass, one missed case forfeits all
efficiency credit. Our tests are finite and pre-layout, so a full pass does not establish equivalence, and wire loads are
statistical, not extracted. Agents may also recall the public upstream designs,
and each result describes a model paired with a single agent harness.
Together, the transcripts, hidden tests, and PPA measurements link the
experiments an agent ran to the behavior and cost of the circuit it built.

\bibliographystyle{IEEEtran}
\bibliography{references}
\end{document}

%% file: analysis/task-scale-macros.tex
\newcommand{\TotalFunctionalCases}{2,808}

%% file: analysis/reward-intervals.tex
\newcommand{\BlackBoxModelDifference}{+0.005}

\newcommand{\GuidedModelDifference}{-0.017}

\newcommand{\GuidanceFunctionalMean}{3.49}
\newcommand{\GuidanceFunctionalLower}{-0.66}
\newcommand{\GuidanceFunctionalUpper}{8.38}
\newcommand{\GuidanceFunctionalTopThree}{3.56}
\newcommand{\GuidanceFunctionalOtherMean}{-0.08}

%% file: analysis/task-table-compact.tex
\begin{table}[t]
\caption{The 18 tasks: reference-core cells (fixed SRAMs excluded), hidden test cases, and their total cycles.}
\label{tab:tasks}
\centering\footnotesize
\setlength{\tabcolsep}{4pt}
\renewcommand{\arraystretch}{1.02}
\begin{tabular}{@{}llrrr@{}}
\toprule Task & Circuit & Cells (k) & Cases & Cycles (k)\\\midrule
\multicolumn{5}{@{}l}{\textit{CPUs and GPUs}}\\
CV32E40P~\cite{gautschi2017ri5cy} & RISC-V core & 61.1 & 396 & 163.8\\
VeeR EH1~\cite{repo-veer} & RISC-V core & 174.5 & 142 & 51.6\\
Vortex~\cite{tine2021vortex} & GPGPU & 648.4 & 163 & 111.0\\
VeriGPU~\cite{repo-verigpu} & GPU & 6.7 & 147 & 924.5\\
\addlinespace[2pt]
\multicolumn{5}{@{}l}{\textit{ML accelerators}}\\
NVDLA-derived~\cite{repo-nvdla} & Conv.\ pipeline & 21.5 & 109 & 132.1\\
RedMulE~\cite{tortorella2023redmule} & Matrix engine & 341.4 & 155 & 31.8\\
OpenTPU~\cite{repo-opentpu} & Systolic TPU & 78.2 & 183 & 74.5\\
ITA~\cite{islamoglu2023ita} & Attention engine & 256.2 & 138 & 50.3\\
\addlinespace[2pt]
\multicolumn{5}{@{}l}{\textit{Media codecs and compression}}\\
JPEG~\cite{repo-jpeg} & Image decoder & 68.6 & 168 & 3,838.1\\
JPEG-LS~\cite{repo-jpegls} & Lossless encoder & 20.3 & 150 & 621.3\\
Gzip~\cite{repo-gzip} & Compressor & 50.6 & 108 & 897.2\\
\addlinespace[2pt]
\multicolumn{5}{@{}l}{\textit{Transform, coding, and cryptography}}\\
FFT~\cite{repo-fft} & 1024-point FFT & 25.8 & 101 & 580.0\\
RS~\cite{repo-rs} & Reed--Solomon dec. & 3.7 & 112 & 886.1\\
Camellia~\cite{repo-camellia} & Block cipher & 44.8 & 125 & 8.8\\
\addlinespace[2pt]
\multicolumn{5}{@{}l}{\textit{Interconnect and memory}}\\
WB DMA~\cite{repo-wbdma} & DMA controller & 53.6 & 158 & 46.1\\
DDR3 AXI~\cite{repo-ddr3} & DRAM controller & 7.1 & 178 & 10,986.1\\
\addlinespace[2pt]
\multicolumn{5}{@{}l}{\textit{I/O and control}}\\
Ethernet~\cite{repo-ethernet} & UDP/IP stack & 10.6 & 169 & 84.3\\
FOC~\cite{repo-foc} & Motor controller & 25.3 & 106 & 17,845.6\\
\bottomrule
\end{tabular}
\end{table}

%% file: analysis/main-results.tex
\begin{table*}[!t]
\centering
\caption{Results (54 trials per row). $r$ and $F$ are means over all trials; PPA ratios and $S$ are geometric means over full passes only. Calls, Probe, and Time are per-trial medians of all agent tool calls (sub-agents included), of those that drive the probe, and of agent working hours; one call may simulate many stimuli of any length, so neither counts waveforms or simulated cycles. Bold marks the highest $r$ and $F$ in each setting.}
\label{tab:main-results}
\small
\setlength{\tabcolsep}{5pt}
\begin{tabular}{@{}lrrrrrrrrrr@{}}
\toprule
& \multicolumn{6}{c}{All 54 trials} & \multicolumn{4}{c}{Full passes only} \\
\cmidrule(lr){2-7}\cmidrule(l){8-11}
Model & $r\uparrow$ & $F\uparrow$ & Full passes & Calls & Probe & Time (h) & $A/A_{\rm ref}\downarrow$ & $P/P_{\rm ref}\downarrow$ & $D/D_{\rm ref}\downarrow$ & $S\uparrow$ \\
\midrule
\multicolumn{11}{@{}l}{\emph{Black-box setting}} \\
GPT-6-Astra & \textbf{0.472} & \textbf{0.759} & 20/54 & 401 & 55 & 5.4 & 1.242 & 1.509 & 1.138 & 0.469 \\
Claude Opus 5.5 & 0.467 & 0.616 & 21/54 & 294 & 49 & 4.5 & 1.101 & 1.138 & 1.048 & 0.761 \\
Kimi K3.1 Pro & 0.115 & 0.210 & 7/54 & 669 & 56 & 5.5 & 2.116 & 3.345 & 1.127 & 0.125 \\
DeepSeek V4.1 Flash & 0.027 & 0.053 & 2/54 & 695 & 104 & 4.8 & 3.940 & 6.184 & 1.121 & 0.037 \\
Qwen4-Max & 0.026 & 0.050 & 1/54 & 345 & 76 & 3.7 & 3.070 & 3.686 & 1.032 & 0.086 \\
GLM-5.3 & 0.006 & 0.012 & 0/54 & 248 & 77 & 4.8 & -- & -- & -- & -- \\
\midrule
\multicolumn{11}{@{}l}{\emph{Function-guided setting}} \\
GPT-6-Astra & 0.481 & \textbf{0.783} & 25/54 & 371 & 60 & 4.5 & 1.581 & 2.090 & 1.148 & 0.264 \\
Claude Opus 5.5 & \textbf{0.499} & 0.682 & 21/54 & 289 & 34 & 4.6 & 1.082 & 1.211 & 1.022 & 0.747 \\
Kimi K3.1 Pro & 0.069 & 0.133 & 2/54 & 723 & 42 & 6.0 & 2.542 & 3.367 & 1.010 & 0.116 \\
DeepSeek V4.1 Flash & 0.091 & 0.178 & 3/54 & 783 & 126 & 5.3 & 2.478 & 5.057 & 1.035 & 0.077 \\
Qwen4-Max & 0.042 & 0.083 & 0/54 & 372 & 66 & 3.2 & -- & -- & -- & -- \\
GLM-5.3 & 0.025 & 0.049 & 0/54 & 289 & 67 & 5.0 & -- & -- & -- & -- \\
\bottomrule
\end{tabular}
\end{table*}